\documentclass[11pt,a4paper]{article}
\usepackage[hyperref]{acl2017}
\usepackage{times}
\usepackage{url}
\usepackage{latexsym}
\usepackage{hyperref}
\usepackage{graphicx}
\usepackage[cmex10]{amsmath}
\usepackage{array}
\usepackage{mdwmath}
\usepackage{mdwtab}
\usepackage{eqparbox}
\usepackage{stfloats}
\usepackage{xcolor}
\usepackage{amssymb}
\usepackage{epsfig}
\usepackage{amssymb}
\usepackage{booktabs}
\usepackage[]{algorithm2e}
\usepackage{floatrow}
\usepackage[font={small}]{caption}

\aclfinalcopy 

\title{Training Language Models Using Target-Propagation}

\author{
 	Sam Wiseman \\
 	Harvard SEAS \\
   	{\small {\tt swiseman@seas.harvard.edu}} \\\And
    Sumit Chopra \\
    Facebook AI Research \\
    {\small {\tt spchopra@fb.com}} \\\And
    Marc'Aurelio Ranzato \\
 	Facebook AI Research \\
 	{\small {\tt ranzato@fb.com}} \\\And
 	Arthur Szlam \\
 	Facebook AI Research \\
 	{\small {\tt aszlam@fb.com}} \\\AND
    Ruoyu Sun \\
 	Facebook AI Research \\
 	{\small {\tt ruoyu@fb.com}} \\\And
    Soumith Chintala \\
    Facebook AI Research \\
    {\small {\tt soumith@fb.com}} \\\And
    Nicolas Vasilache \\
    Facebook AI Research \\
    {\small {\tt ntv@fb.com}}
    }

\date{}

\begin{document}
\maketitle
\begin{abstract}
While Truncated Back-Propagation through Time (BPTT) is the most popular approach to training Recurrent Neural Networks (RNNs), it suffers from being inherently sequential (making parallelization difficult) and from truncating gradient flow between distant time-steps.
We investigate whether Target Propagation (TPROP) style approaches can address these shortcomings.
Unfortunately, extensive experiments suggest that TPROP generally underperforms BPTT, and we end with an analysis of this phenomenon, and suggestions for future work.
\end{abstract}

\section{Introduction}
\label{sec:introduction}
Modern RNNs are trained almost exclusively using truncated 
Back-Propagation Through Time (BPTT)~\cite{elman1990,werbos1990backpropagation,mikolov-2010}. 
Despite its popularity, BPTT training has two major drawbacks, namely, that it is inherently sequential (and thus difficult to parallelize), and that it truncates the number of time-steps over which gradient information can propagate.  

Inspired by the recent reports of success of Target-Propagation (TPROP) in training non-recurrent deep networks~\cite{carreira-perpinan14,lee2015difference,taylor16admm}, we 
explore training RNNs with TPROP, an idea that has been suggested repeatedly in the literature~\cite{lecun1986,lecun1988,mirowski_lecun2009}. TPROP can be understood as a generalization of backpropagation, where neural networks are trained by providing local targets for each hidden unit. Such approaches have been motivated by appealing to biological plausibility, numerical stability, computational parallelizability, and its conduciveness to constrained training~\cite{lecun1986,lecun1987,lecun1988,krogh1989cost,bengio14,carreira-perpinan14,lee2015difference}.

Concretely, we treat the hidden states of an RNN 
as free variables, which are optimized independently, but encouraged to be predictable from previous hidden states. Formulating the model in this way offers an approach to avoiding the sequential 
nature of BPTT training, by allowing for optimization over the parameters 
and all hidden states simultaneously for the entire data set, which is easily parallelized.

We extensively evaluate applying TPROP to train language models, and we find: (1) that batch TPROP is as effective as Batch BPTT in minimizing training loss, but they both fail to generalize well; 
(2) that mini-batch TPROP achieves comparable generalization performance to BPTT only in the limit when TPROP reduces to 
mere BPTT; (3) that the relatively unconstrained nature of the hidden state optimization appears to be responsible for the performance degradation.



\subsection{Review: Truncated BPTT} 
\label{subsec:bptt}
Letting $x_t$ represent the input at time $t$, an RNN produces hidden states $h_t$ and predictions $\hat{y}_t$ at each time-step using the following recurrence

\vspace*{-4mm}
\begin{align}
h_t &= g(x_t, h_{t-1}), \label{eq:standard-recur}\\
\hat{y}_t &= f(h_t), 
\end{align}

\vspace*{-0mm}
\noindent where $g$ and $f$ are non-linear, parametric functions of their inputs. 
For instance, an Elman RNN for language modeling might be specified by defining $g(x_t, h_{t-1}) = \sigma(W_x x_t + W_h h_{t-1})$, and $f(h_t) = \mathrm{softmax}(W_y h_t)$, where $\sigma$ denotes the logistic sigmoid function. 

It is typical to apply RNNs to sequence-prediction problems that have a
well-defined loss $\ell(\hat{y_t}, y_{t})$ at every
time-step, where $y_t$ is the desired output at the $t$'th step.
In the case of language modeling, $\ell$ is the cross-entropy loss, where the number of classes is equal to the number of words in the vocabulary, and $y_t = x_{t+1}$.

\begin{figure}[!t]
\includegraphics[width=1\linewidth]{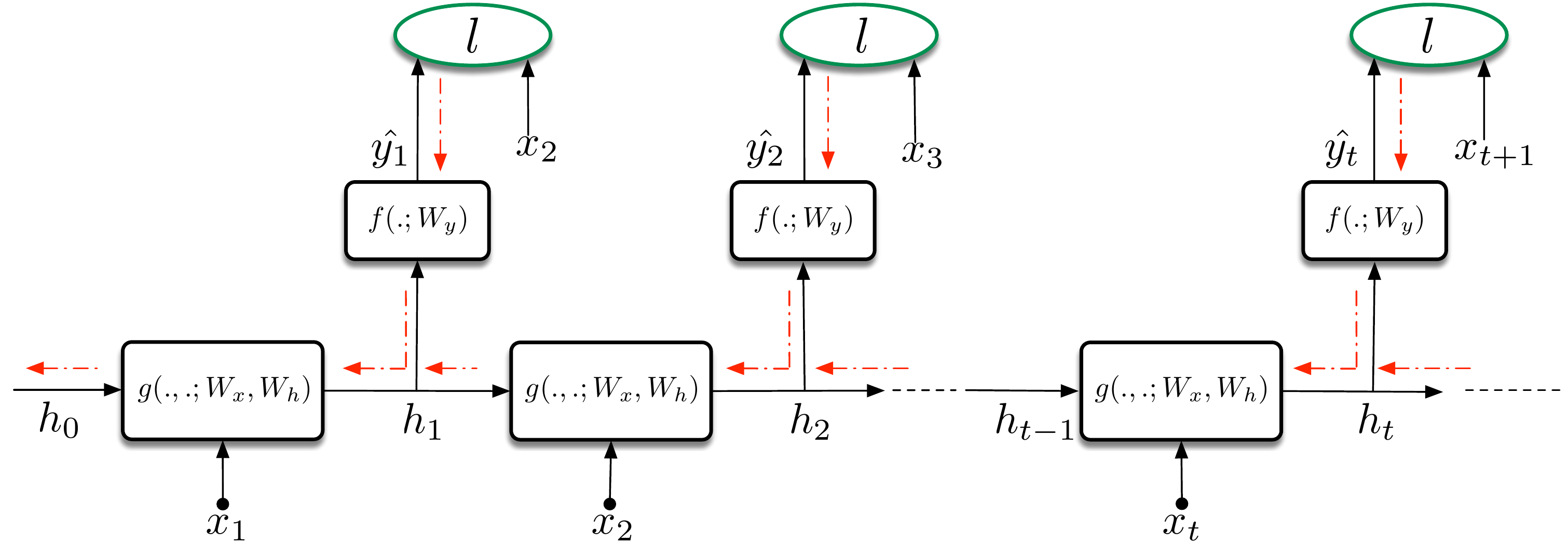}
\caption{An RNN unrolled for $t$ time-steps during training. Black arrows indicate the forward direction of the unrolled recurrence, and the red arrows the direction of the gradients.}
\label{fig:rnn_train}
\end{figure}

Given $\ell$, as well as a dataset consisting of an input sequence
$x_1,\ldots,x_T$ and desired output sequence $y_1,\ldots,y_T$, it is possible
to obtain a loss for the entire dataset in terms of any parameters $\theta$
(and an initial hidden state $h_0$)
by unrolling the RNN in time, as depicted in 
Figure~\ref{fig:rnn_train}.
In particular, we have

\vspace*{-4mm}
\begin{eqnarray}
\mathcal{L}(\theta) & = & 
    \sum_{t=1}^T \ell(f(h_t), x_{t+1}) \nonumber \\
	& = & \ell(f(g(x_1,h_{0})), x_{2}) \label{eq:unrolled} \\
	  & &    + \ell(f(g(x_2, g(x_1,h_{0})), x_{3})  
			  + \ldots \nonumber
\end{eqnarray}

\noindent Typically the loss in Equation~\ref{eq:unrolled} is not optimized in batch. 
Rather, the sequence is unrolled for a window of $K$ steps, 
and a gradient step is taken to minimize the loss on those $K$ steps. After updating parameters, the minimization continues by unrolling the next (non-overlapping) window of $K$ steps, initializing $h_0$ to the value of the $K$'th hidden state in the previous window, minimizing the loss on this new window, and continuing through the dataset in this way. 
Crucially, each window is considered to be independent of all previous ones, and so no gradients are calculated with respect to time-steps in previous windows. Going forward, we will refer to the loss induced by this windowing and truncation as the ``Truncated BPTT Loss.''

\vspace*{-1mm}
\paragraph{Issues with Truncated BPTT}
The Truncated BPTT Loss and associated minimization procedure suffers from two problems. 
First, the training algorithm is 
inherently sequential: one cannot process two consecutive spans of size $K$ in parallel, since the processing of a $K$-window depends on having processed the previous $K$-window, which makes parallelization impossible. Second, the hard truncation of the gradients beyond $K$ steps 
makes it difficult for the network to capture long-term 
dependencies in the data.

\section{Target Propagation}
\label{sec:model}
In order to address the issues associated with the truncated BPTT
algorithm, we propose to train RNNs with a slightly modified loss. 
In particular, rather than unrolling the losses $\ell$ over time-steps (as in Equation~\ref{eq:unrolled}), which has the effect of instantiating the $h_t$ only implicitly, we instead treat the $h_t$ as explicit variables to be optimized over. 
In order to maintain the recurrent 
property of the model, however, we add additional constraint terms to the loss, which encourage adjacent hidden states to approximately obey the parametric recurrence.

\begin{figure}[!t]
\includegraphics[width=1\linewidth]{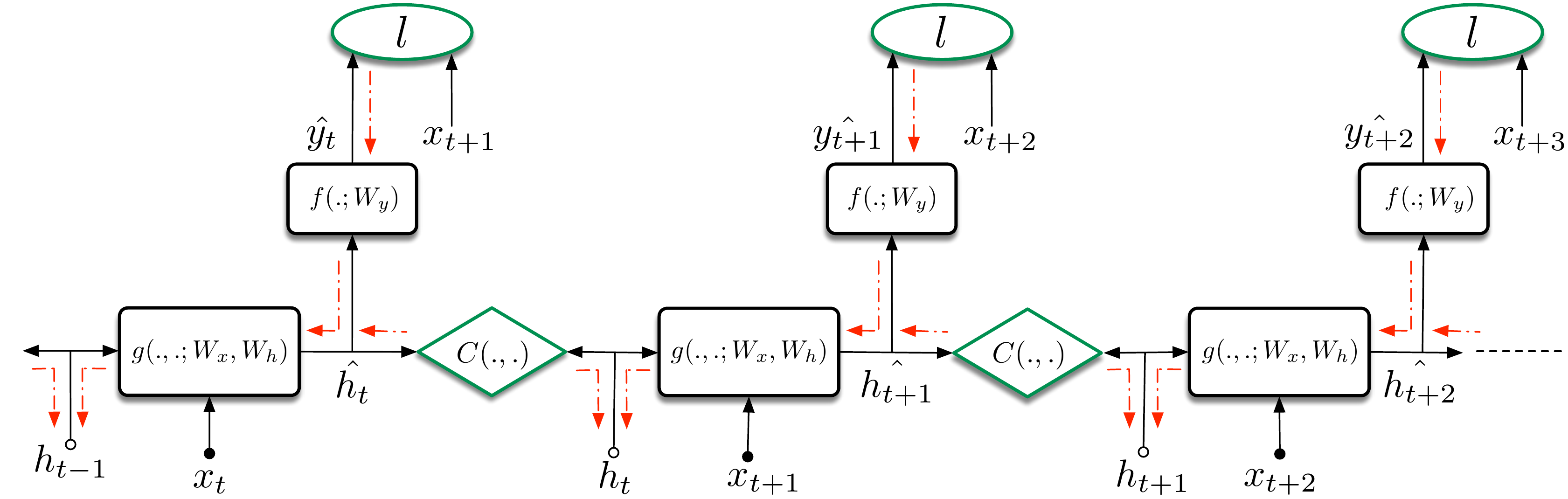}
\caption{Schematic of TPROP training. The diamonds 
depict the constraint functions, which take $\hat{h}_t$ and $h_t$ as input. Hollow circles indicate free 
variables, and solid circles indicate fixed, observed variables. 
The free variables $h_t$s receives gradients from both 
the constraint and the recurrent unit.}
\label{fig:rnn_tprop}
\end{figure}

More concretely, we define $\hat{h}_t$ to be the predicted hidden state at time $t$, as follows:
\begin{eqnarray}
\hat{h}_t & = & g(x_t, h_{t-1}) \label{eq:tprop-recur}\\
\hat{y_t} & = & f(\hat{h}_t).
\end{eqnarray}
Note that Equation~\eqref{eq:tprop-recur} uses the independent variable $h_{t-1}$ on the right-hand-side; otherwise it is equivalent to Equation~\eqref{eq:standard-recur}.
Using $\mathcal{H}$ to refer to the set of all $h_t$, we then modify the loss to be: 
\begin{equation}
\mathcal{L}(\mathcal{H}, \theta) = \sum_{t=1}^T \ell(f(\hat{h_t}), x_{t+1}) + \lambda C(\hat{h_t}, h_t), 
\label{eq:tprop_loss}
\end{equation}
where $C$ is a penalty or constraint function (e.g., an $L_2$ penalty) introduced
to force the predicted hidden state $\hat{h_t}$ at time step $t$ to 
be close to the free hidden state $h_t$, and $\lambda$ is a coefficient 
whose value governs how strictly we wish to enforce the constraint.
This approach is referred to as Target-Propagation because the $h_t$ and the $\hat{h}_t$ serve as targets for each other during optimization. See Figure~\ref{fig:rnn_tprop} for reference. 

\subsection{The Blocked Target-Propagation Algorithm (BTPROP)}
We also consider a generalization of the TPROP loss, where instead of making each $h_t$ a free variable, we only have a free variable every $B$ time steps. The remaining hidden states are defined implicitly through the recurrence~\eqref{eq:standard-recur}, which further constrains the model during training. We refer to the sub-sequence of length $B$ consisting of time-steps beginning with a free variable and ending before the next free variable as a ``block.'' Note that with block size $B=1$ we recover the TPROP formulation in the previous subsection. 


\paragraph{Benefits of BTPROP} We emphasize that the BTPROP loss offers an approach to addressing the issues with the Truncated BPTT Loss identified in Section~\ref{subsec:bptt}.
First, the independence of non-boundary time steps in different blocks suggests that training may be efficiently parallelized by distributing a large number of (contiguous) blocks to each machine in a cluster, which would necessitate inter-machine communication only for the small number of $h_t$ that border blocks on a different machine. We note, however, that since parameters $\theta$ are shared across time-steps (and must therefore be synchronized between machines), for such a scheme to offer a speed-up it must also be the case that $\mathcal{H}$ and $\theta$ can be optimized independently, and that the $\mathcal{H}$-optimization results in faster convergence of the $\theta$-optimization.

To address the second issue with BPTT identified in Section~\ref{subsec:bptt}, we note that the penalty terms $C$ encourage the final hidden state of a block to be close to the initial hidden state of the subsequent block, which should allow for the capturing of dependencies between multiple blocks during training. Also note that by restricting the block size we can restrict the temporal dependence of the loss on any given $h_t$, thereby mitigating the vanishing or exploding gradient problem.

Finally, we note that we expect BTPROP to offer an advantage over TPROP (with $B$ = 1), since it leads to a more constrained optimization problem (with fewer variables), and because it allows for intra-block BPTT, which has become a relatively mature technology.


\section{Training}
\label{sec:training}
The loss in Equation~\ref{eq:tprop_loss} can be seen as deriving from an equality-constrained formulation of the cross-entropy loss. In particular, if we introduce equality constraints between each $h_t$ and $\hat{h}_t$, as well as a dual variable $u_t$ for each constraint, the Lagrangian can be written (up to a constant) as
\begin{align} 
&\mathcal{L}_{\mathrm{aug}}(\mathcal{H}, \theta, \mathcal{U}) = \nonumber \\ 
&\; \sum_{t=1}^T \ell(f(\hat{h_t}), x_{i+1}) + \frac{\lambda}{2} ||h_t - \hat{h}_t + u_t||^2, \label{eq:admm_loss}
\end{align}
which is in the form of Equation~\ref{eq:tprop_loss} with $C$ defined as $C(\hat{h_t}, h_t) = \frac{1}{2} ||h_t - \hat{h}_t + u_t||^2$.

It is natural to minimize the now-unconstrained loss \eqref{eq:admm_loss} using an Alternating Direction Method of Multipliers (ADMM) style approach~\cite{glowinski1975approximation,gabay1976dual,boydadmm}, which results in the following meta-algorithm:
\begin{enumerate}
\item[1.] Minimize $\mathcal{L}_{\mathrm{aug}}$ with respect to $\mathcal{H}$
\item[2.] Minimize $\mathcal{L}_{\mathrm{aug}}$ with respect to $\theta$
\item[3.] Update duals: $u_t \gets u_t + \alpha_u \nabla_{u_t} \mathcal{L}_{\mathrm{aug}}$.
\end{enumerate}

\noindent We note that for many choices of RNN architecture it will be impossible to perform either the $\theta$-minimization or the $\mathcal{H}$-minimization analytically.\footnote{Exceptions to this include various forms of multiplicative RNNs~\cite{wumult} with no non-linearity. In this case, alternating minimizations can be carried out with least squares. We experimented along these lines, but found such methods to underperform gradient based approaches, which are also much more flexible.} As such, we simply use gradient-based algorithms for a fixed number of steps, and we find that it is not necessary (and generally not advisable) to minimize until convergence.

We also consider two alternatives to the above algorithm. The first, which we refer to as the Penalty Method (PM)~\cite{courant1943variational,nocedal2006numerical}, is identical to ADMM, except that it avoids the use of dual variables entirely, and so skips step 3. The second, the Augmented Langrangian Method (ALM)~\cite{hestenes1969multiplier,powell1967method,nocedal2006numerical}, minimizes jointly over $\mathcal{H},\theta$ before updating the dual variables, effectively merging steps 1. and 2. We found that ADMM outperforms PM, and very minimally outperforms ALM as well, and so we report ADMM results in what follows.

\section{Experiments}
\label{sec:experiments}
We run word-level language modeling experiments on the Penn Tree Bank (PTB)~\cite{marcus1993building} and Text8 datasets.\footnote{http://mattmahoney.net/dc/textdata} 
For all experiments we use single-layer Gated Recurrent Unit (GRU) RNNs~\cite{cho2014learning} (without Dropout~\cite{srivastava2014dropout}), and we use Adagrad~\cite{duchi2011adaptive} for optimization. 

We report the perplexity obtained on the validation dataset after freezing the final parameters $\theta$ obtained during the alternating optimization process; no optimization is done at test time. (Preliminary experiments suggested that slightly better results can be obtained by optimizing over \textit{past} $h_t$ variables at test-time, though we did not pursue this direction due to its inefficiency.)



\subsection{Results}
\label{sec:results}
We report our main results in Table~\ref{tab:main}, where we compare BTPROP validation perplexity performance for various block-sizes and various numbers of $\mathcal{H}$ steps with BPTT. We report only validation numbers because training performance between BTPROP and BPTT was generally comparable.

Starting from the left portion of Table~\ref{tab:main}, we see that BTPROP is roughly comparable to BPTT for bigger $B$. Importantly, however, BTPROP performance does not tend to improve with additional $\mathcal{H}$-steps. This is disappointing because it can be shown that BTPROP with a single $\mathcal{H}$-step is essentially equivalent to BPTT; see the Supplemental Material for a more rigorous formulation and proof. Furthermore, it is clear from the table that both Batch BTPROP and Batch BPTT are significantly outperformed by their minibatch counterparts, presumably due to the regularization effect of updating parameters after seeing only a subset of the data (see \newcite{keskar2016large} for a discussion of this phenomenon). This suggests that even if Batch BTPROP were more easily parallelized, it would not be worth the decreased generalization. 


\setlength{\tabcolsep}{4pt}
\begin{table*}[t]
   \centering
   \begin{tabular}{lccc@{\hskip 0.7cm}ccc@{\hskip 0.7cm}ccc}
     \toprule
      & \multicolumn{3}{c}{Batch PTB PPL} & \multicolumn{3}{c}{Minibatch PTB PPL} & \multicolumn{3}{c}{Minibatch Text8 PPL} \\[2pt] 
           & $B=5$ & $B=10$ & $B=20$ & $B=5$ & $B=10$ & $B=20$ & $B=5$ & $B=10$ & $B=20$\\ 
     \cmidrule{2-10}
      $\mathcal{H}$-steps = 1 & 192.00  & 201.08 & 180.08 & 137.27  & 130.93 & 127.43 & 242.02  & 220.11 & 205.67\\     
     $\mathcal{H}$-steps = 2 & 204.14  & 189.70 & 179.42 & 137.31  & 131.04 & 127.91 & 242.23  & 216.54 & 206.54 \\
     $\mathcal{H}$-steps = 5 & 226.16  & 184.60  & 188.99 & 137.37  & 132.95 & 128.56 & 241.47  & 216.87 & 204.60 \\
     \midrule
     BPTT ($K {=} B$) & 188.31 & 182.58 & 183.91 & 135.73 & 129.50 & 128.16 &  229.22 & 209.87 & 201.07\\
   \bottomrule
   \end{tabular}
   \caption{Validation perplexities for Batch PTB (left), Minibatch PTB (middle), and Minibatch Text8 (right), varying the number of $\mathcal{H}$-optimization steps and $B$. We compare with BPTT performance when window-size $K = B$.}
   \label{tab:main}
\end{table*}

We now consider the right two portions of Table~\ref{tab:main}, which involve Minibatch BTPROP. We note that applying BTPROP in a minibatch setting is not straightforward, since the penalty term $C(\hat{h_t}, h_t)$ will now involve hidden states $h_t$ that have not been updated since the previous epoch, and which therefore may not provide reasonable targets. We address this by initializing the hidden states $h_t$ in the current minibatch to the $\hat{h_t}$ given by the current parameters before starting the $\mathcal{H}$-optimization. We also emphasize that in the minibatch setting there is little hope for a parallelization gain; it is possible, however, that the BTPROP loss will still allow the model to benefit from training with longer-range dependencies. Unfortunately, the results again suggest that BTPROP gains little over BPTT, and that moreover additional $\mathcal{H}$-steps are not in general beneficial. 

\subsection{BTPROP as $L_2$ Regularizer}
Additional experiments summarized in Table~\ref{tab:ablation} suggest that the small BTPROP performance gains in the Minibatch setting (see Table~\ref{tab:main}) are attributable to the regularization effect of the $L_2$ penalty in~\eqref{eq:admm_loss}, rather than training with longer dependencies. There, we compare the Validation PPL obtained on Minibatch PTB with $\mathcal{H}$-steps = 1 and $B$ = 20 with doing no $\mathcal{H}$-optimization (which improves (i.e., lowers)) perplexity, and with doing no $\mathcal{H}$-optimization but also setting $\lambda=0$ (which significantly hurts (i.e., increases) perplexity).

\begin{table}[t]
   \centering
   \begin{tabular}{lc}
     \toprule
       & $\Delta$ PPL \\
      \midrule
      $\mathcal{H}$-steps = 0 & -0.40 \\     
     $\mathcal{H}$-steps = 0 \& $\lambda$ = 0 & +2.36 \\
   \bottomrule
   \end{tabular}
   \caption{Changes in PPL, using $\mathcal{H}$-steps = 1, B = 20 on Minibatch PTB as a baseline. Setting $\mathcal{H}$-steps = 0 \textit{improves} PPL as long as $\lambda > 0$.}
   \label{tab:ablation}
\end{table}

\subsection{A Possible Explanation and Future Work}
\begin{figure}
\hspace*{-0.5cm} \includegraphics[scale=0.43]{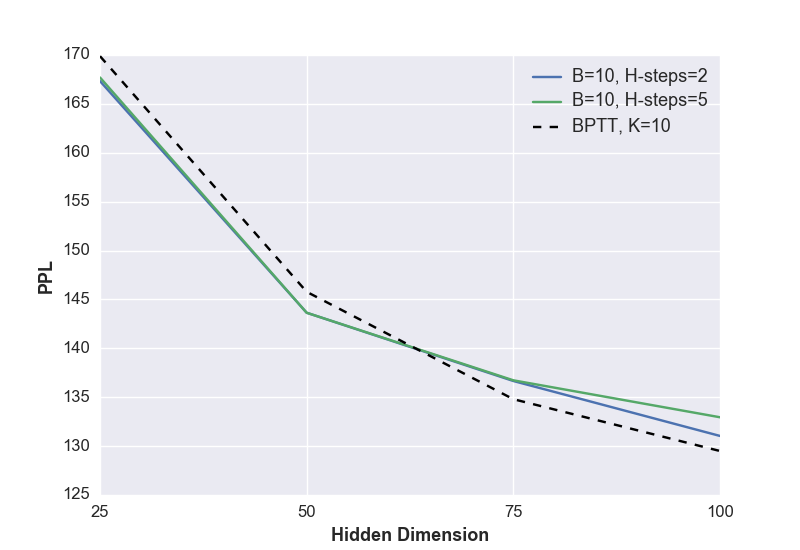}
\caption{Validation perplexities obtained from BTPROP ($B=10$) and BPTT ($K=10$) as the dimension of the hidden states increases. BPTT's improvement increases with the dimensionality.}
\label{fig:diffppl}
\end{figure}

One explanation for the $\mathcal{H}$-optimization hurting performance is that its relatively unconstrained nature may allow for finding hidden states that decrease training loss without leading to parameters that generalize. Indeed, many of the reported successes of TPROP-style training have involved very constrained problems, such as those with binary hidden-state constraints~\cite{perpinan15} or sparsity constraints~\cite{koray-cvpr-09}. To test this hypothesis, in Figure~\ref{fig:diffppl} we plot the perplexity obtained from BTPROP with $B$ = 10 (for both 2 and 5 $\mathcal{H}$-optimization steps) as well as the perplexity obtained from BPTT with $K$ = 10, as the dimensionality of the hidden state increases. We see that BPTT actually underperforms BTPROP for small hidden states, but improves relatively as the hidden state size increases, lending support to our hypothesis. A similar pattern emerges when comparing BTPROP with $B$ = 5 and BPTT with $K$ = 5. This suggests that getting BTPROP to work with larger hidden states may involve finding additional approaches to constraining the $\mathcal{H}$-optimization, such as by, for instance, penalizing the KL divergence between the distributions $\mathrm{softmax}(W_y \hat{h}_t)$ and $\mathrm{softmax}(W_y h_t)$, which we leave to future work.

\bibliography{references}
\bibliographystyle{acl_natbib}

\appendix

\newpage
\section{Supplemental Material}


\subsection{Connection between BPTT and TPROP}
Here we show that when using the Penalty Method loss (or, equivalently, the ADMM loss while keeping the dual variables $u_t$ set to 0), doing BTPROP with $\mathcal{H}$-steps = 1 and $\theta$-steps = 1 results in gradients with respect to $\theta$ that are equal to the gradients with respect to $\theta$ obtained from BPTT (up to a constant factor), if: (1) the $\mathcal{H}$ are initialized such that $h_t = g_{\theta}(x_t, h_{t-1})$; (2) the $h_t$ are updated with vanilla gradient descent.


More formally, let $\ell(g_{\theta}(x_t, h_{t-1}), y_t)$ be a per-time-step loss, where $\theta$ are the parameters of $g$. Also define
\begin{align} \label{eq:laug}
\ell_{\mathrm{pm}} = \ell(h_t, y_t) + \frac{\lambda}{2} ||g_{\theta}(x_t, h_{t-1})-h_t||^2.
\end{align} 
We show that if we initialize $h_t = g_{\theta}(x_t, h_{t-1})$ then we have
\begin{align} \label{eq:target_eq2}
\frac{\partial \, \ell_{\mathrm{pm}}(\tilde{h}_t, y_t)}{\partial \, \theta} = \eta \lambda \frac{\partial \, \ell(g_{\theta}(x_t, h_{t-1}), y_t)}{\partial \, \theta},
\end{align}
where we have defined
\begin{align} \label{eq:htilde2}
\tilde{h}_t = h_t - \eta \frac{\partial \, \ell_{\mathrm{pm}}(h_t, y_t)}{\partial \, h_t} .
\end{align}

\noindent From the definition of $\ell_{\mathrm{pm}}$, we have
\begin{align*}
\frac{\partial \, \ell_{\mathrm{pm}}(h_t, y_t)}{\partial \, h_t} &= \frac{\partial \, \ell(h_t, y_t)}{\partial \, h_t} + \lambda (h_t -  g_{\theta}(x_t, h_{t-1})) \\
&= \frac{\partial \, l(h_t, y_t)}{\partial \, h_t},
\end{align*}
where the last line follows from the fact that $h_t = g_{\theta}(x_t, h_{t-1})$.  Substituting into \eqref{eq:htilde2} then gives
\begin{align*}
\tilde{h}_t = h_t - \eta \frac{\partial \, \ell(h_t, y_t)}{\partial \, h_t}.
\end{align*}

\noindent Now, from the definition of $\ell_{\mathrm{pm}}$ and the chain rule, we can write the left hand side of~\eqref{eq:target_eq2} as
\begin{align}
&\frac{\partial \, \ell_{\mathrm{pm}}(\tilde{h}_t, y_t)}{\partial \, \theta} \nonumber\\
&= \frac{\partial \, g_{\theta}(x_t, h_{t-1})}{\partial \, \theta} \lambda \Big( g_{\theta}(x_t, h_{t-1}) - \tilde{h}_t \Big) \nonumber\\
&= \frac{\partial \, g_{\theta}(x_t, h_{t-1})}{\partial \, \theta} \lambda \Big( g_{\theta}(x_t, h_{t-1}) -  \nonumber\\
& \qquad h_t  + \eta \frac{\partial \, \ell(h_t, y_t)}{\partial \, h_t} \Big) \nonumber\\
&= \frac{\partial \, g_{\theta}(x_t, h_{t-1})}{\partial \, \theta} \lambda \left(\eta \frac{\partial \, \ell(h_t, y_t)}{\partial \, h_t} \right), \label{eq:lhs2}
\end{align}
where the last line again follows from the fact that $h_t = g_{\theta}(x_t, h_{t-1})$. Since we can also write the right hand side of ~\eqref{eq:target_eq2} as
\begin{align*}
\frac{\partial \, \ell(g_{\theta}(x_t, h_{t-1}), y_t)}{\partial \, \theta} &= \frac{\partial \, g_{\theta}(x_t, h_{t-1})}{\partial \, \theta} \frac{\partial \, \ell(h_t, y_t)}{\partial \, g_{\theta}(x_t, h_{t-1})} \\
&= \frac{\partial \, g_{\theta}(x_t, h_{t-1})}{\partial \, \theta}  \frac{\partial \, \ell(h_t, y_t)}{\partial \, h_t},
\end{align*}
we obtain the equality as desired.

\subsection{Hyperparameter Grid Used in Experiments}
See Table~\ref{tab:hypers}.

\begin{table}
   \centering
   \begin{tabular}{lc}
     \toprule
     $\lambda$ & \{1, 0.1, 0.01\} \\    
     $u_t$ learning rate & \{1, 0.1, 0.01\}  \\
     $\mathcal{H}$-optimization learning rate & \{0.1, 0.01, 0.001\}  \\
     $\mathcal{\theta}$-optimization learning rate & \{0.1, 0.01, 0.001\}  \\   $\mathcal{\theta}$-optimization steps & \{1\} \\    
     $\mathcal{H}$-optimization steps & \{1, 2, 5\} \\
   \bottomrule
   \end{tabular}
   \caption{Grid of hyperparameters used in experiments.}
   \label{tab:hypers}
\end{table}

\subsection{Increasing Mini-Batch Size with BPTT}
An obvious approach to speed up training consists of increasing the mini-batch size of SGD. Unfortunately, the results in table~\ref{tab:bigminibatch} suggest that performance deteriorates on both training and test sets whenever the minibatch size is big enough
to leverage parallel computation (mini-batch size of size 4096 and above). These findings are in line with the finding in Section~\ref{sec:results} that both batch BPTT and batch BTPROP are significantly inferior to their mini-batch counterparts.

\begin{table}[htbp]
\begin{center}
\begin{tabular}{l||c|c|c|c|c}
WIKI & 64 & 256 & 1024 & 4096 & 16000 \\
\hline
\hline
training & 29.7 & 33 & 40.7 & 81 & 1070 \\
\hline
test & 47.7	& 49.1	& 52 & 78.7 & 938 
\end{tabular}

\begin{tabular}{l||c|c|c|c|c}
TEXT8 & 64 & 256 & 1024 & 4096 & 16000 \\
\hline
\hline
training & 1.35 &1.37
 & 1.36 & 1.38 & 1.46 \\
\hline
test & 1.39 & 1.41 & 1.42 & 1.45 & 1.47
\end{tabular}
\end{center}
\caption{Varying the mini-batch size (columns) when training with BPTT using the wiki-large$\_$103  dataset at the word level, and the text8 dataset at the character level. In the wiki-large$\_$103, we used a two layer LSTM with 512 hidden units unrolled for 30 steps. In the text8 dataset, we used a one layer LSTM with 1024 hidden units unrolled for 75 time steps. In both cases, increasing the mini-batch
size increases perplexity on both training and test sets.}
\label{tab:bigminibatch}
\end{table}

\subsection{Code}
Code implementing our experiments can be found at \url{https://github.com/facebookresearch/TPRNN}

\end{document}